\pdfoutput=1
\documentclass[11pt]{article}

\usepackage[review]{emnlp2021}

\usepackage{times}
\usepackage{latexsym}

\usepackage[T1]{fontenc}

\usepackage[utf8]{inputenc}

\usepackage{microtype}

\usepackage{comment}
\usepackage{hyperref}
\usepackage{amsmath}
\usepackage{nccmath}
\usepackage{graphicx}
\usepackage{float}

\title{Voice2Action: Language Models as Agent for \\ Efficient Real-Time Interaction in Virtual Reality}

\author{Yang Su \\
  Cornell Tech \\
  \texttt{ys724@cornell.edu} \\}

\begin{document}
\maketitle
\begin{abstract}
Large Language Models (LLMs) are trained and aligned to follow natural language instructions with only a handful of examples, and they are prompted as task-driven autonomous agents to adapt to various sources of execution environments. However, deploying agent LLMs in virtual reality (VR) has been challenging due to the lack of efficiency in online interactions and the complex manipulation categories in 3D environments. In this work, we propose Voice2Action, a framework that hierarchically analyzes customized voice signals and textual commands through action and entity extraction and divides the execution tasks into canonical interaction subsets in real time with error prevention from environment feedback. Experiment results in an urban engineering VR environment with synthetic instruction data show that Voice2Action can perform more efficiently and accurately than approaches without optimizations. 
\end{abstract}

\section{Introduction}

Large Language Models (LLMs) have demonstrated impressive zero-shot and few-shot learning abilities in natural language understanding and generation \cite{brown2020language}. With human alignments like reinforcement learning from human feedback (RLHF), these models become better at following human instructions \cite{ouyang2022training}; with instruction prompting and providing external resources \cite{nakano2021webgpt}, they can be used as agents to autonomously choose tools \cite{schick2023toolformer}, communicate with other agents \cite{shen2023hugginggpt}, and show superior ability in decision-making and task execution.

However, the seamless integration of these models within VR has remained a challenging frontier, hindered by efficiency, accuracy, and the complexities associated with interactions and manipulations in 3D spaces. Firstly, as a simulated three-dimensional interaction environment that mimics the real world, the VR environment has enormous possibilities in the way that the user can interact with entities (objects in the virtual scene) and manipulate their properties; secondly, the game engines that execute the user instructions has a pre-defined set of atomic operations for entity attribute modifications, causing it non-trivial to map or classify the user instruction to the most proper configuration in the engine; lastly, the accuracy of VR hardware (i.e., the voice recognition SDK, \href{https://wit.ai/}{Wit.ai}) and the efficiency in 3D graphics rendering (i.e., the uv rendering pipeline) limits the number of operations we can perform while not exceeding user's comfortable response time to receive the feedback of the executed tasks.

In this paper, we focus on two main challenges for deploying agent LLMs in VR: efficiency and accuracy. While improving and balancing these metrics, we plan to define how agent LLMs operate within the virtual environments, and then build an interactive tool to provide users with a more practical experience in developing their customized virtual scene. Hence, we propose the Voice2Action framework, created upon a rich taxonomy of text input commands, ranging from simple object selection and state manipulation to more complex operations involving animation, scripted sequences, and environment configuration modification. By hierarchical instruction prompting and entity extraction, Voice2Action can accurately interpret users' textual instructions by incorporating environmental feedback.

To provide empirical validation, we conduct experiments and ablation studies in an urban city planning virtual environment. We build a synthetic dataset generated by the \emph{text-davinci-003} model from \href{https://platform.openai.com/docs/models/model-endpoint-compatibility}{OpenAI API} with the self-instruct \cite{wang2022self} framework, where we use a pre-defined canonical instruction subset as the seed tasks, and manually filter out the unsatisfactory generated instruction-execution pair. The results indicate a marked increase and well-balanced execution efficiency and accuracy.

To summarize, our contributions to operating agent LLMs in VR are:

\begin{enumerate}
    \item We define a hierarchical set of canonical instructions in VR for language models to perform decision-making actions.
    \item We improve efficiency and accuracy by incorporating different-purposed LLMs to divide and conquer the execution tasks, and error prevention from environment feedback.
    \item We build and open source the Voice2Action framework\footnote{Code is available at \url{https://github.com/yang-su2000/VR-Multimodal-Interaction}} to stimulate further research for applying agent LLMs in customizable 3D interactive environments.
\end{enumerate}

\section{Related Work}

\textbf{Language models for decision-making.} Large language models' robust text understanding ability enables them to make accurate decisions and perform action execution. By using external resources such as browsing the web \cite{nakano2021webgpt}, using tools by deciding API calls \cite{schick2023toolformer}, or connecting different LLMs and summarizing their responses \cite{shen2023hugginggpt}, language models can be used as controllers for any task executions. For more complex tasks, LLMs are also prompted to plan, self-criticize by reasoning and acting \cite{yao2022react}, and store history in memory with external databases, as shown by \href{https://github.com/yoheinakajima/babyagi}{BabyAGI} and \href{https://github.com/hwchase17/langchain}{LangChain}. Many of these methods rely on extensive human feedback from policy learning or freely generate inefficient long reasoning chains to reach the final decision. Voice2Action uses a divide-and-conquer framework that hierarchically extracts and performs the action so that an unnecessary decision-making trajectory can be pruned before the generation chain is finished, which makes it both efficient and relies more on the environment than human feedback.

\textbf{Language models in an interactive environment.} Deploying LLMs in an interactive environment has been studied mainly in robotics planning \cite{huang2022inner} and closed-loop systems \cite{yao2022react}, where real-time environment changes are generally ignored. In the VR environment, efficiency and accuracy affected by real-time graphics rendering, state changes, and human intervention are essential components. To the best of our knowledge, Voice2Action is the first to tackle the problem of using LLM for decision-making in a real-time VR environment.

\begin{figure*}[ht]
\centering
\centering
    \includegraphics[scale=0.47]{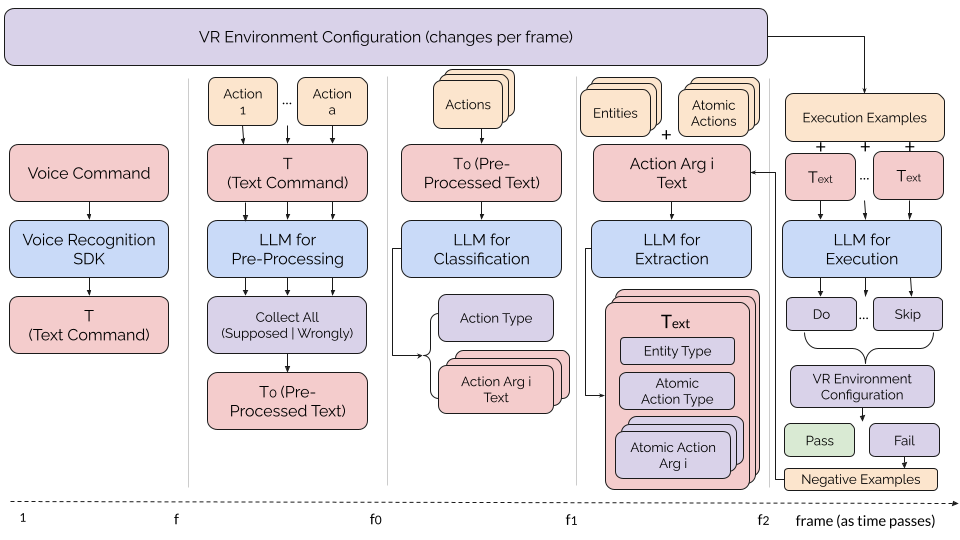}
    \caption{\label{fig:1}
An overview of Voice2Action. The process consists of 5 steps that take place sequentially, where red refers to \textcolor{red}{audio or text data}, orange refers to \textcolor{orange}{textual prompts or examples}, blue refers to \textcolor{blue}{audio recognition or language models}, purple refers to \textcolor{violet}{structured arguments or configurations}, and green refers to that \textcolor{green}{the program successfully executes}.
}
\end{figure*}

\section{Background}

Consider the general setup of a user interacting within the VR environment (Figure \ref{fig:1}). Due to the uncomfortableness of keyboard typing (text commands) in virtual scenes, most users begin their commands through voice signals. From frame $1$ to $f$ (usually 60 frames per second), the user initiates a voice command with input voice sequence $V=\{V_1,..., V_f\}$, which goes through a black box voice recognition hardware SDK that sequentially produces a text sequence $T=\{T_1, ..., T_t\}$ by the end of the last frame $f$, and the environment changes along the frame $E=\{E_1,..., E_f\}$. On average, $t\approx 0.04f$. We do not restrict the format of $T$ generated from the user's commands. However, $T$ is highly inaccurate, and it is necessary to perform additional pre-processing with \textbf{agent LLM for preprocessing} from frame $f+1$ to $f_0$. In particular, 

\begin{equation}
\begin{split}
T_0 &= LLM_{pre}(T, S_{action})
\end{split}
\end{equation}

where $S_{action}$ is a dictionary of $(\textcolor{red}{action\ type}, \textcolor{blue}{action\ args})$ pairs.

From frame $f_0+1$ to $f_1$, an \textbf{agent LLM for classification} takes in $T_0$ and identifies the action category and the textual commands associated with the action as a formalized template $T_1$. In particular, 

\begin{equation}
\begin{split}
T_1 &= LLM_{cls}(T_0, S_{action}) = \\ 
& \textcolor{red}{action\ type:} T_{action\ type}, \\ 
& \textcolor{blue}{action\ arg_1:} T_{action\ arg_1}, ..., \\
& \textcolor{blue}{action\ arg_n: } T_{action\ arg_n}
\end{split}
\end{equation}

where each $T_{action\ arg_i}$ are filtered natural language sequences from $T_0$, and $n$ is the number of argument for $\textcolor{red}{action\ type}$.

From frame $f_1+1$ to $f_2$, an \textbf{agent LLM for extraction} recognizes $T_{action\ type}$ and extracts the action and entity properties from each $T_{action\ arg_i}$ as a formalized template $T_2$. In particular, 

\begin{equation}
\begin{split}
T_2 &= concat(i=1\to n: T_{ext}) \\
T_{ext} &= LLM_{ext}(T_{action\ arg_i}, S_{entity}, S_{atomic\ action}) \\
&=  \textcolor{red}{entity:} T_{entity\ type}, \\
& \textcolor{red}{atomic\ action\ type:} T_{atomic\ action\ type}, \\
& \textcolor{blue}{atomic\ action\ arg_1:} T_{atomic\ action\ arg_1}, ..., \\
& \textcolor{blue}{atomic\ action\ arg_m:} T_{atomic\ action\ arg_m}
\end{split}
\end{equation}

where $S_{entity}$ is the collection of available entity (object) types in the virtual environment, and $S_{atomic\ action}$ is the collection of available built-in atomic functions in the game engine. $T_{atomic\ action\ arg_j}$ could deviate from natural languages (i.e., they can be non-text form), as long as they match the correct argument types defined in its atomic action type, and $m$ is the number of arguments for $\textcolor{red}{atomic\ action\ type}$.

The environment $E=\{E_1,...E_{f_2}\}$ changes while the above operations are performed, which is treated as the environment feedback. From frame $f_2+1$ to $f_3$, an \textbf{agent LLM for execution} takes in $T_{ext}$ with its specified arguments, decides the order and number of actions to perform based on the atomic action list and environment feedback, and generates a sequence of actions $A=\{A_1..., A_a\}$ with length $a=|A|$. In particular,

\begin{equation}
\begin{split}
A &= concat(i=1\to a: A_{exe}) \\
A_{exe} &= LLM_{exe}(T_{ext}, E) = respond(T_{ext})
\end{split}
\end{equation}

where $respond\in\{do, skip\}$. Each $A_{exe}$ happens exactly on some frame from $f_2+1$ to $f_3$. To reduce execution difficulty, we do not consider the correlation of consecutive voice commands, or the environment changes after frame $f_2$, and only focus on the efficiency and accuracy of the agent LLMs generated output $A$. 
We also fix all $A_{exe}$ to be performed at frame $f_3$, which makes the execution \textit{time-invariant}, i.e., no actions would affect the outcomes of other actions. Time-invariant actions include many canonical manipulation tasks in the virtual environment, such as object selection, positioning, and static property modifications.

\section{Method}

We propose an efficient autonomous interaction framework in the VR environment that hierarchically uses language models as agents to extract and execute user commands with error prevention from the environment feedback. We break down the problem into two stages: (1) how an "action" is defined in VR; and (2) how our method executes actions in VR.

Action in the VR environment is defined by its task space, task parameters, and design dimension \cite{laviola20173d}. The task space can be treated as the category of our dataset, which outlines the scope of interaction and the configuration of the VR environment. In our experiment, the task space is an urban engineering virtual scene where the users want to manipulate the building layouts for city planning. The users are usually equipped with means to do object selection and mesh manipulation, as implemented in \href{https://www.arkio.is/}{Arkio}, a popular building sketching VR tool. Hence, $S_{action}={(select, select\ args), (mesh, mesh\ args)}$. 

Task parameters define the specific goals or objectives that the user is supposed to achieve within the task space. In the urban engineering example, task parameters might include locating specific items, navigating through the city, and moving or altering the properties of the items. i.e., $select\ args=\{object\ type,\ location,\ direction,\ ...\}$. 

Design dimensions refer to the specific VR elements and mechanisms that are used to facilitate the accomplishment of the task.  In our example, the design dimension would involve how buildings, roads, and vehicles are oriented. i.e., $S_{entity}=\{building,\ road,\ vehicle\}$ and $S_{atomic\ action}=\{range(start, end),\ locate(x,y,z), ...\}$.

With these parameters defined, our method consists of the following 4 modules to create a hierarchical NLP system for interpreting and executing commands in a VR environment.

\subsection{Voice to Text Pre-processing} The voice commands go through a black-box speech recognition hardware, and the produced text is often misaligned with their pronunciation (i.e., trees $\to$ we, building $\to$ beauty). To clean and pre-process the text command, we first use an \textbf{agent LLM for pre-processing} with instruction prompting to reconstruct the tokens to the closest pronunciation in our task space. We sample the output from $m_{pre}=|S_{action}|$ LLMs, where each of them is provided $k_{pre}$ example commands with a particular action type. The outputs are prompted and structured as $(supposed, wrongly\ pronounced)$ token pairs. We collect and weight these pairs by dividing the number of occurrences of the $supposed$ tokens in our instruction dataset, and select the top $\alpha=25\%$ $wrongly\ pronounced$ token to replace. This module vastly improves execution accuracy by mapping wrongly pronounced tokens to more frequent tokens in the dataset.

\begin{table*}[ht]
\centering
\begin{tabular}{lll}
\hline
\textbf{Notation} & \textbf{Example Input/Output} & \textbf{Example Prompts}\\
\hline
$T$ & select the highest & N/A\\
& beauty on mean sea & \\
$T_0$ & select the highest & find likely misspelled words in \{action type\} \\
& building on main street & \\
$T_{action\ type}$ & select & "select" refers to \{explanation\}\\
$T_{entity\ type}$ & building & extract entity type from $S_{entity}$\\ 
$T_{select\ arg_1}$ & height & if applicable, extract superlative degree of "building" \\
$T_{select\ arg_2}$ & main street & if applicable, extract location of "building" \\
$T_{atomic\ type}$ & scale getter & "scale getter" refers to \{function documentation\} \\
$T_{atomic\ action\ arg_1}$ & y: inf & if applicable, execute action \\
& & "getter(y=inf)" on entity "building" \\
\hline
\end{tabular}
\caption{\label{tab:1}
Examples of input and outputs of the Voice2Action framework, voice inputs are omitted. When the game engine starts, available action and entity types will be loaded and filled in \{...\}, along with the documentation of atomic functions.
}
\end{table*}

\begin{table*}[ht]
\centering
\begin{tabular}{lllllll}
\hline
\textbf{Model} & $N_0$ & $N_1$ & $N_2$ & $N_3$ & $N_{trial}$ & $N_{token}$ \\
\hline
LLM-Exe & 0 & 0 & 0 & 368 & 5.4 & 1987 \\
LLM-Pre-Exe & 152 & 0 & 0 & 355 & 2.9 & 1182 \\
LLM-Pre-Ext-Exe & 152 & 0 & 402 & 133 & 1.3 & 848 \\ 
Voice2Action & 152 & 92 & 285 & 140 & \textbf{1.2} & \textbf{754} \\
\hline
\end{tabular}
\caption{Efficiency results on different model architectures by averaging over the synthetic dataset with 100 data samples. We see that the full model uses much fewer trials and generates fewer tokens than previous baselines.}
\label{tab:2}
\end{table*}

\subsection{Text Property Classification} The pre-processed text $T$ needs to be categorized into one of several predefined action categories. Based on 3D interaction principles \cite{laviola20173d}, these categories could include 
\begin{itemize}
    \item static entity property modification actions: including object selection, mesh manipulation, creation, and static state changes.
    \item dynamic entity property modification actions: object movements, mesh transformation (animations), dynamic state changes, and scripted sequences (house starts firing).
    \item environment property modification actions: environment control, physics property alternations, and graphics effect changes.
\end{itemize}
For efficiency consideration, we limit $S_{action}$ to include only selection and mesh manipulation in the urban engineering scenario. We provide the set of action categories to the \textbf{agent LLM for classification} and their corresponding action types, along with $k_{cls}$ few-shot examples, chosen upon the complexities of the task parameters. We only use 1 LLM as this sub-task is simpler, and the output is structured as $T_1$.

\subsection{Entity and Action Extraction} For each formalized $T_1$ template consisting of the action type and its arguments in natural language format, we use an \textbf{agent LLM for extraction} to decompose the $(entity, action)$ pair into atomic function calls in the VR system. This is usually infeasible with the number of atomic actions defined in any game engine, so we restrict the set to the canonical manipulation tasks defined in \cite{laviola20173d}. It consists of atomic actions used for object selection and mesh manipulations, which fits our design dimension of urban engineering. We use $m_{ext}=|S_{atomic\ action}|$ LLM to do the extraction for each $T_1$ with $k_{ext}$ examples. As $m_{ext}$ gets larger, this operation would be costly. We thus improve efficiency by pre-calculating the vector embeddings of the atomic actions' textual description and only extracting the closest atomic action that matches each $T_{action\ arg_i}$ on the fly.

\subsection{Command Execution} We have now gathered both the structured $action\ type$ in $T_1$ and $(entity, atomic\ action)$ pair in $T_2$, the \textbf{agent LLM for execution} needs to determine the order of execution and filter out unnecessary actions with feedback from the environment. We sample $m_{exe}$ LLM to generate the execution orders and prompt them to freely filter out the actions that they consider unnecessary; then we put each of them in a simulated environment to execute the commands and receive the $feedback\in\{pass,\ fail\}$. If it fails, we go back to the extraction step with an added negative example from the error message until it passes. Finally, we pick the execution LLM that takes the shortest time (which usually generates shorter texts) and successfully accomplishes the task. In practice, we stop other LLMs from running once the task is successfully accomplished and modify the environment state to the designated successful configuration, which significantly improves accuracy by preventing error occurrences from the environment. An example flow of the framework is listed in Table \ref{tab:1}.

\section{Experimental Setup}

We generate a synthetic dataset with 100 data samples. Each sample contains a textual command $T_0$. We create this dataset with some modifications to the self-instruct framework \cite{wang2022self}. Starting with $S_{action}, S_{entity}$, and $S_{atomic\ action}$, we randomly sample one action type, $d_{ent}=Uniform(1,3)$ entity types, and $d_{atom}=Uniform(2,10)$ atomic actions, and provide them to the \emph{gpt-4} module from OpenAI API along with their textual description (API documentation), and some example use cases (manually written by human). We re-sample 10 times and ask the model to generate 10 $T_0$ samples each round. $d_{ent}$ and $d_{atom}$ are chosen by a user study to mimic the number of operations that an urban engineer would do in a virtual city planning scenario. After all $T_0$ is collected, we manually filter out the ones that are not satisfactory enough, which empirically turns out to be less than 10\%. We re-sample a few more times and keep 100 of them in the experiment. The dataset is small because to run each sample, we need to manually speak out the sentence of $T_0$ to go through the voice recognition SDK and wait for all agent LLMs and the environment $E$ to finish its frame and provide feedback. We manually evaluate the correctness of each $(T_0, E)$ pair by investigating the environment and the usage of entities and atomic actions. On average, running and evaluating each pair costs 20 seconds and 0.015\$.

For the agent LLMs, we use \emph{text-davinci-003}. For the pre-processing agent, we set the temperature to 0.9 to encourage diverse pronunciation interpretation; for the other agents, we set the temperature to 0 to discourage randomness. 80\% confidence and 512 maximum generation length are set for all agents. For few-shot examples, $k_{pre}=k_{ext}=k_{exe}=3, k_{cls}=|S_{action}|=2$ are provided for the agent LLMs.

We compare the following baselines and improved methods. In practice, we run them all paralleled in the simulated VR environment and evaluate them one by one.
\begin{itemize}
    \item LLM-Exe: the model that takes in $T$, with added prompts for choosing between extracting and execution, additionally provided with $k_{ext}$ examples.
    \item LLM-Pre-Exe: the model that takes in pre-processed $T_0$ and call LLM-Exe. This model mimics the decision-making LLMs proposed in related works, which lets the model choose between planning, reasoning, and executing.
    \item LLM-Pre-Ext-Exe: the full model but treats $T_{action\ arg_i}$ as the pre-processed $T_0$. This model essentially puts the task of classification to the extraction and execution module.
    \item Voice2Action (LLM-Pre-Cls-Ext-Exe): the full model.
\end{itemize}

\begin{figure}[ht]
\centering
\centering
    \includegraphics[scale=0.5]{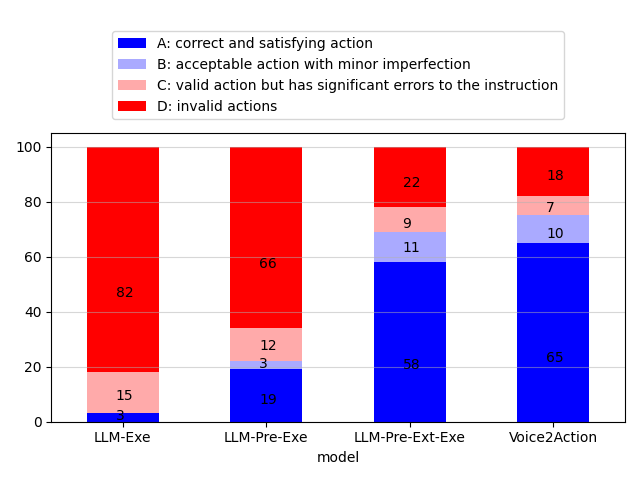}
    \caption{\label{fig:2}
Accuracy results on different model architectures. Human evaluators rate the model's action by 4 levels based on the VR environment feedback.
}
\end{figure}

\section{Results}

The accuracy is measured by the performance of $A_{exe}$. We rate $A_{exe}$ by 4 levels suggested by self-instruct \cite{wang2022self}, as shown in Figure \ref{fig:2}. The first 3 levels correspond to the $pass$ state from environment feedback, and the last level D corresponds to the $fail$ state, i.e., an error has occurred in the game engine, usually due to illegal parse of the action arguments.

The efficiency is measured by the number of tokens $N_{token}$. In particular, 
$$N_{token}=N_0 + N_1 + N_{trial}*(N_2+N_3)$$
$$N_i=|T_i|+|prompt(T_i)| $$
where $N_i$ is the total token length from both the input tokens from instruction prompting and the intermediate tokens generated from agent LLMs, and $N_{trival}$ is the number of runs for the model to receive the $pass$ flag from the environment. Note that a $pass$ simply means that the code can execute, which does not indicate correctness. Higher $N_{token}$ means that the number of rendered frames $f_3$ in the environment caused by the performed action is higher, which is of critical importance to the user's VR experience. The details are in Table \ref{tab:2}.

\textbf{Analysis} For accuracy comparison, we see that the model without voice command pre-processing can barely execute any action correctly, because most wrongly-generated tokens are pronunciation-wise similar to the supposed token rather than semantically similar. The model with both the extraction and execution LLMs has much better decision-making accuracy, compared to the ones with only one of these LLMs, because the extraction LLM "divides" the tasks into atomic blocks for the execution LLM to "conquer" it, which can be verified from the generated token length in Table \ref{tab:2}. The classification LLM makes little improvement as we only evaluate 2 action types, which is a simpler task that can be approachable without further "dividing". 

For efficiency comparison, we see that models without "divide-and-conquer" have significantly larger $N_{trial}$, which suggests that the model cannot successfully map the commands to the atomic actions defined in the game engine. We also see that the full model with the added classification LLM makes the generated token length $N_i$ for each component more evenly distributed, which implies the effectiveness of the hierarchical framework in Voice2Action.

\section{Conclusion}

We build a framework to enable efficient real-time interaction in virtual reality built upon agent language models, called Voice2Action. By hierarchically dividing the users' commands into class-specific categories and matching them with atomic function calls in the game engine, this system shows exceptional interaction efficiency with minimal loss of execution accuracy. By incorporating environment changes and user feedback with models from other modalities, the system also demonstrates the potential to be applicable to more complex environments, where multi-turn and multi-agent interaction can be further developed. We open-source the framework and leave this as future work for the community to work on.

\bibliography{emnlp2021}
\bibliographystyle{acl_natbib}

\end{document}